\PassOptionsToPackage{dvipsnames}{xcolor}
\PassOptionsToPackage{pdfpagelabels=false}{hyperref} 

\documentclass{IOS-Book-Article}

\usepackage{mathptmx}
\usepackage[T1]{fontenc}
\usepackage{soul}\setuldepth{article}
\usepackage{booktabs}
\usepackage{amssymb}
\usepackage{colortbl}
\usepackage[relative,overlay]{textpos}
\usepackage{fancyvrb}
\usepackage{orcidlink}
\RequirePackage[skins]{tcolorbox}
\newtcolorbox{custombox}[1]{
	colback=gray!10,
	colframe=gray!70,
	left=1mm,
	right=1mm,
	top=1mm,
	bottom=1mm,
	fonttitle=\bfseries,
	arc=0mm,
	leftrule=1mm,
	rightrule=0mm,
	toprule=0mm,
	bottomrule=0mm,
	notitle,
	before=\par\smallskip\noindent,
	before upper={\textbf{#1 } },
}
\def\hb{\hbox to 11.5 cm{}}

\definecolor{theftinashop}{RGB}{234,150,163}
\definecolor{anotherobject}{RGB}{228,148,106}
\definecolor{atwork}{RGB}{194,154,75}
\definecolor{pickpocketing}{RGB}{166,159,70}

\definecolor{visit}{RGB}{133,164,65}
\definecolor{openaccess}{RGB}{72,176,94}
\definecolor{car}{RGB}{74,173,143}
\definecolor{car-light}{RGB}{237,247,243}
\definecolor{residence}{RGB}{75,171,164}

\definecolor{vehicle}{RGB}{78,171,184}
\definecolor{bicycle}{RGB}{80,169,214}
\definecolor{energy}{RGB}{157,170,233}
\definecolor{metals}{RGB}{199,159,233}

\definecolor{cellars}{RGB}{231,138,224}
\definecolor{others}{RGB}{232,145,193}

\definecolor{negativehl}{RGB}{255,194,179}
\definecolor{positivehl}{RGB}{218,255,179}

\newcommand{\hln}[2][negativehl]{ {\sethlcolor{#1} \hl{#2}} }
\newcommand{\hlp}[2][positivehl]{ {\sethlcolor{#1} \hl{#2}} }

\usepackage[strict]{changepage}

\usepackage{framed}

\definecolor{formalshade}{rgb}{0.95,0.95,1}

\newenvironment{formal}{%
  \MakeFramed{\advance\hsize-\width\FrameRestore}%
  \noindent\hspace{-4.55pt}
  \begin{adjustwidth}{}{7pt}%
  \vspace{2pt}\vspace{2pt}%
}
{%
  \vspace{2pt}\end{adjustwidth}\endMakeFramed%
}

\begin{document}

\pagestyle{headings}
\def\thepage{}
\begin{frontmatter}              

\title{Using Large Language Models\\ to Support Thematic Analysis\\ in Empirical Legal Studies}

\runningauthor{J. Dr\'apal, H. Westermann, and J. Savelka}
\runningtitle{Using LLMs to Support Thematic Analysis in ELS}

\author[A,B]{\fnms{Jakub} \snm{Dr\'apal}\orcidlink{0000-0001-9455-9013}%
\thanks{Corresponding Author: Jakub Dr\'apal,  drapalja@prf.cuni.cz. Work supported by Czech Grant Agency project ``Sentencing disparities in the post-communist continental legal systems'' n. 19-15077S.}},
\author[C]{\fnms{Hannes} \snm{Westermann}\orcidlink{0000-0002-4527-7316}},
and
\author[D]{\fnms{Jaromir} \snm{Savelka}\orcidlink{0000-0002-3674-5456}
}

\address[A]{Institute of State and Law of the Czech Academy of Sciences, Czechia}
\address[B]{Institute of Criminal Law and Criminology, Leiden University, the Netherlands}
\address[C]{Cyberjustice Laboratory, Universit\'e de Montr\'eal, Canada}
\address[D]{School of Computer Science, Carnegie Mellon University, USA}

\begin{abstract}
Thematic analysis and other variants of inductive coding are widely used qualitative analytic methods within empirical legal studies (ELS). We propose a novel framework facilitating effective collaboration of a legal expert with a large language model (LLM) for generating initial codes (phase 2 of thematic analysis), searching for themes (phase 3), and classifying the data in terms of the themes (to kick-start phase 4). We employed the framework for an analysis of a dataset~($n=785$) of facts descriptions from criminal court opinions regarding thefts. The goal of the analysis was to discover classes of typical thefts. Our results show that the LLM, namely OpenAI's GPT-4, generated reasonable initial codes, and it was capable of improving the quality of the codes based on expert feedback. They also suggest that the model performed well in zero-shot classification of facts descriptions in terms of the themes. Finally, the themes autonomously discovered by the LLM appear to map fairly well to the themes arrived at by legal experts. These findings can be leveraged by legal researchers to guide their decisions in integrating LLMs into their thematic analyses, as well as other inductive coding projects.
\end{abstract}

\begin{keyword}
Thematic analysis\sep empirical legal studies\sep criminal law\sep large language models\sep generative pre-trained transformers\sep GPT-4
\end{keyword}
\end{frontmatter}

\section{Introduction}

\begin{figure}
    \flushright
    \includegraphics[width=\textwidth]{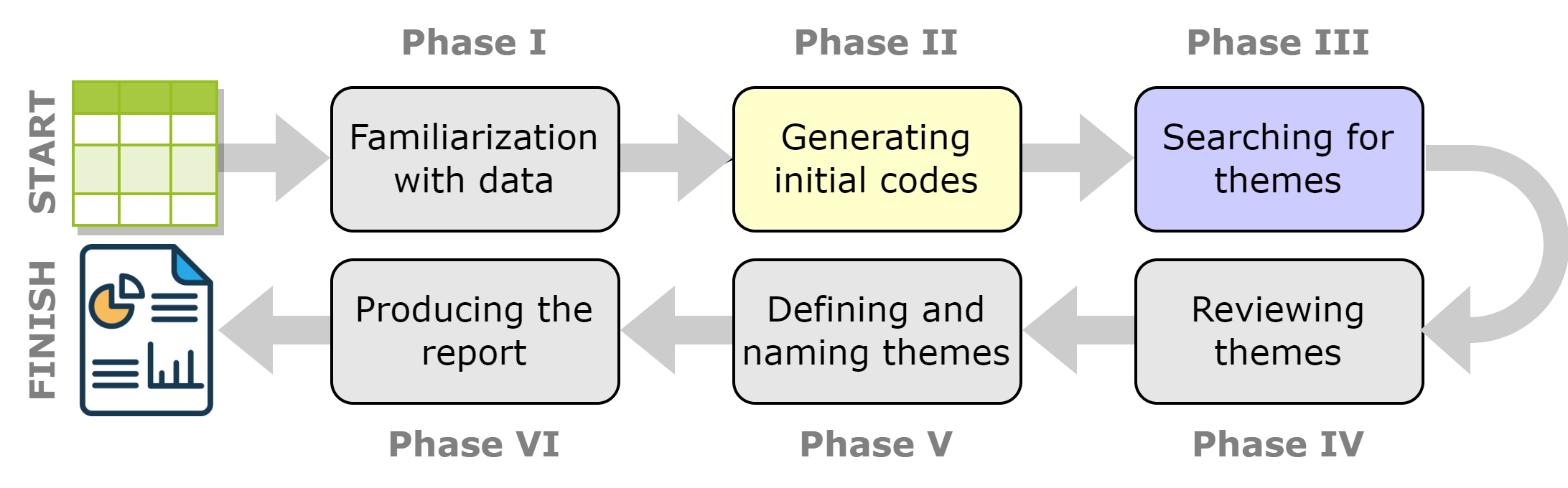}
    \caption{The six phases of thematic analysis, starting from the raw data and finishing with the scholarly report. This paper focuses on supporting phase II (yellow) and phase III (blue) of the analysis with LLMs.}
    \label{fig:thematic_analysis}
\end{figure}

Empirical legal studies (ELS) is an approach to the study of law through empirical methods typical of economics, psychology, and sociology. Since law is a heavily text-based discipline ELS frequently focuses on text analytic methods, including deductive and inductive coding. Deductive coding focuses on applying a fixed set of codes to a dataset, whereas inductive coding leads to a simultaneous discovery of the codes from the data and their application. While investigations into various methods to support deductive coding have attracted much recent attention in AI \& Law \cite{westermann2019computer,branting2020scalable,westermann2020sentence} very few studies focused on inductive coding~\cite{salaun2022tenants}. One popular inductive coding method is ``thematic analysis'' \cite{braun2006using}. While rarely explicitly acknowledged thematic analysis 
is widely used in ELS. Figure \ref{fig:thematic_analysis} shows the six phases of the analysis, starting from the raw data and finishing with the scholarly report on the studied phenomena. We propose a novel LLM-powered framework to support a subject matter expert in performing phases~2 and 3 of the analysis.

We employed the proposed framework in a thematic analysis of criminal courts' opinions, focused on the criminal offense of theft in Czechia. Criminal offense categories~(e.g., theft, murder) are usually defined in statutory law while the individual criminal acts are described by courts when they apply the law to factual circumstances of the cases. An important question in criminal law and criminology is what behaviors are \textit{actually} criminalized and whether it is done appropriately. Neither the statutory definitions of the offenses (they are too general and not ``sociologically relevant'' \cite{kitsuse1963note}) nor the descriptions of the factual circumstances from cases (too specific) can answer the question. To get insight into what behavior is criminalized and how effectively, it is necessary to identify shared features of criminal acts, generalize them into ``typical crimes'' \cite{hornle2013moderate,lappi2001sentencing} and arriving at behavioral-based categories \cite{unodc2015international,national2016modernizing}. This is akin to performing thematic analysis. While important, such analysis is an expensive and time-consuming endeavour. Hence, a (semi-)automated approach would be useful.

To assess the capability of a state-of-the-art LLM (GPT-4) to support selected stages of the thematic analysis, we investigated the following research questions: 

\begin{enumerate}
    \item[(RQ1)] How successfully can the LLM perform initial coding of the data?
    \item[(RQ2)] To what degree can a subject matter expert improve the quality of the initial codes via natural language feedback?
    \item[(RQ3)] How successfully can the LLM predict themes for the analyzed data points?
    \item[(RQ4)] How successfully can the LLM autonomously discover themes and associate them with the analyzed data?
\end{enumerate}

\section{Related Work}
There have been several studies exploring the use of \emph{LLMs in thematic analysis}. De Paoli evaluated to what extent GPT-3.5 can carry out a full-blown thematic analysis of semi-structured interviews, finding that the LLM was indeed able to perform some of the steps while also cautioning about the methodological implications of using the approach~\cite{de2023can}. Gao et al. developed a collaborative coding platform powered by GPT-3.5 that provides code and code group suggestions to support the process of defining a codebook \cite{gao2023collabcoder}. Gamieldien et al. used GPT-3.5 to generate codes for automatically clustered comments, finding that the produced codes were granular but not coherent, as similar clusters were assigned very different names \cite{gamieldien2023advancing}.

There is a long tradition of studies identifying \emph{patterns in criminal justice data}, including those focused on offense categories. The studies typically employed content analysis together with approaches such as factor, latent profile or cluster analyses. Santtila et al. identified 14 types of burglaries from the descriptions of crime scene behavior~\cite{santtila2004predicting}. Higgs et al. collated descriptions of 700 sexual murderers to describe the overall patterns and motives underlying the offense \cite{higgs2017sexual}. Canter et al. performed a thematic classification of stranger rapes \cite{canter2003differentiating}. G\v{r}ivna and Dr\'{a}pal focused on criminal offenses involving computer data and systems (cybercrime) in the Czech Republic, identifying the most frequent types of such criminal behavior \cite{gvrivna2019attacks}. 

There is a similar tradition focused on \emph{discovering stereotypical patterns in court opinions} in AI \& Law. Ashley identified factors from trade secret law through reading cases and doctrine \cite{ashley1991reasoning}. Similar analysis was performed by Gray et al. to discover typical factors of suspicion considered in auto stop cases \cite{gray2022toward}. Westermann et al. used the grounded theory approach (a close kin of thematic analysis) to discover relevant factors considered by judges in certain types of landlord-tenant disputes \cite{westermann2019using}. Notably, Salaun et al. used a topic modelling approach in the same domain to identify the factors automatically, finding that 33\% of the  discovered topics were relevant \cite{salaun2022tenants}. To our best knowledge, that study is the state-of-the-art attempt on inductive coding of legal texts. Our work differs by subscribing to a well-established framework (i.e., thematic analysis) and the use of GPT-4 that enables a subject matter expert to drive and influence the analysis through specified research questions and instructions.

There were multiple proposals of \emph{frameworks focused on supporting legal experts} in deductive coding of legal texts. Branting et al. proposed manual annotation of factors in a small number of cases, which were then projected across a much larger dataset \cite{branting2020scalable}. Westermann et al. described an approach where legal experts formulated sets of complex search terms (as classifiers) based on constantly updated dataset statistics \cite{westermann2019computer}. Westermann et al. also proposed a framework utilizing sentence embeddings and similarity retrieval to support annotators in annotating legal documents \cite{westermann2020sentence}. Recently, Savelka et al. explored performing annotations with LLMs (GPT-3.5 and GPT-4) in zero-shot settings by providing  the model with excerpts from annotation guidelines \cite{savelka2023unlocking,savelka2023can,savelka2023unreasonable}. In this paper, we perform deductive coding when predicting the themes for the case facts descriptions (RQ3) as one of the steps in predominantly inductive coding-based analysis.

\section{Dataset}
\label{sec:dataset}
\begin{figure}
\flushleft
\scriptsize
\setlength{\tabcolsep}{3pt}
\begin{tabular}{llll}
\fcolorbox{black}{theftinashop}{\rule{0pt}{3pt}\rule{3pt}{0pt}}~Theft in a shop &  \fcolorbox{black}{visit}{\rule{0pt}{3pt}\rule{3pt}{0pt}}~Theft during a visit &  \fcolorbox{black}{vehicle}{\rule{0pt}{3pt}\rule{3pt}{0pt}}~Theft of motor veh. &  \fcolorbox{black}{cellars}{\rule{0pt}{3pt}\rule{3pt}{0pt}}~Robbing of cellars\\
\fcolorbox{black}{anotherobject}{\rule{0pt}{3pt}\rule{3pt}{0pt}}~Break. into another obj. &  \fcolorbox{black}{openaccess}{\rule{0pt}{3pt}\rule{3pt}{0pt}}~Open-access place theft &  \fcolorbox{black}{bicycle}{\rule{0pt}{3pt}\rule{3pt}{0pt}}~Bicycle theft &  \fcolorbox{black}{others}{\rule{0pt}{3pt}\rule{3pt}{0pt}}~Others\\
\fcolorbox{black}{atwork}{\rule{0pt}{3pt}\rule{3pt}{0pt}}~Theft at work &  \fcolorbox{black}{car}{\rule{0pt}{3pt}\rule{3pt}{0pt}}~Theft from a car  &  \fcolorbox{black}{energy}{\rule{0pt}{3pt}\rule{3pt}{0pt}}~Theft of energy\\
\fcolorbox{black}{pickpocketing}{\rule{0pt}{3pt}\rule{3pt}{0pt}}~Pickpocketing &  \fcolorbox{black}{residence}{\rule{0pt}{3pt}\rule{3pt}{0pt}}~Breaking into a residence&  \fcolorbox{black}{metals}{\rule{0pt}{3pt}\rule{3pt}{0pt}}~Theft of metals\\
\end{tabular}
\flushright
\includegraphics[width=.49\textwidth]{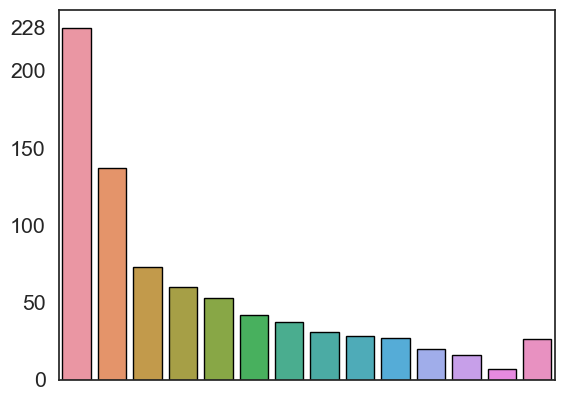}
\begin{textblock*}{6.2cm}(0.05cm,-4.65cm)
\begin{formal}
\scriptsize
At an undetermined time between 18:00 on May 12, 2017 and 06:00 on May 13, 2017, at the parked delivery vehicle branded Peugeot Boxer, an unknown individual used an unidentified object to pry open the locks of the driver's door, the passenger door, and then the cargo space. The individual \textbf{entered the vehicle and stole from it} a car radio, a demolition hammer, an electric saw, a drill, and other work tools, all valued at 8,700 CZK [...] By damaging the door lock, he caused damage worth 3,500 CZK. The stolen items were sold to unknown persons.
\end{formal}
\end{textblock*}
\caption{The categories from the theft types dataset (shown at the top) and their distribution (right). An example of case facts description from the \emph{theft from a car} category is shown on the left.}
\label{fig:dataset}
\end{figure}

In our experiments, we used a dataset of 785 facts descriptions from cases of Czech courts decided in 2017. From the Prosecution Service, we received 834 cases  that found an adult defendant guilty of theft. In Czechia, theft also includes burglary and pick-pocketing.\footnote{ICCS codes 0501 and 0502 except for 0502212 \cite{unodc2015international}.} We slightly over-represented the most serious offenses to ensure a sufficient number of cases in the dataset. We removed 49 cases from the dataset because they were used in a pilot study or due to them containing errors. We extracted text describing the facts. Each extracted text was anonymized and shortened or partially re-written if necessary.
The resulting text snippets range from 73 to 29,695 characters in length (1Q 447, median 782, 3Q 1,462 characters). Figure \ref{fig:dataset} shows an example (automated translation).

A group of three law students under the supervision of one of the authors of this paper manually conducted an unstructured variant of thematic analysis.\footnote{We did not rigorously adhere to the process described in \cite{braun2006using}.} The group arrived at 14 high-level themes focused on modus operandi and target of committed thefts~(Figure \ref{fig:dataset}). For each facts description a single theme was independently chosen by two students according to specified rules. The disagreements were resolved by one of the students following careful re-reading of the case. The distribution of the themes over the 785 facts descriptions included in the dataset is presented in Figure \ref{fig:dataset}. The \emph{theft in a shop}~(29.0\%) and \emph{breaking into another object} (17.5\%) are the most prevalent themes.

\section{Proposed Framework}

\begin{figure}
    \includegraphics[width=\textwidth]{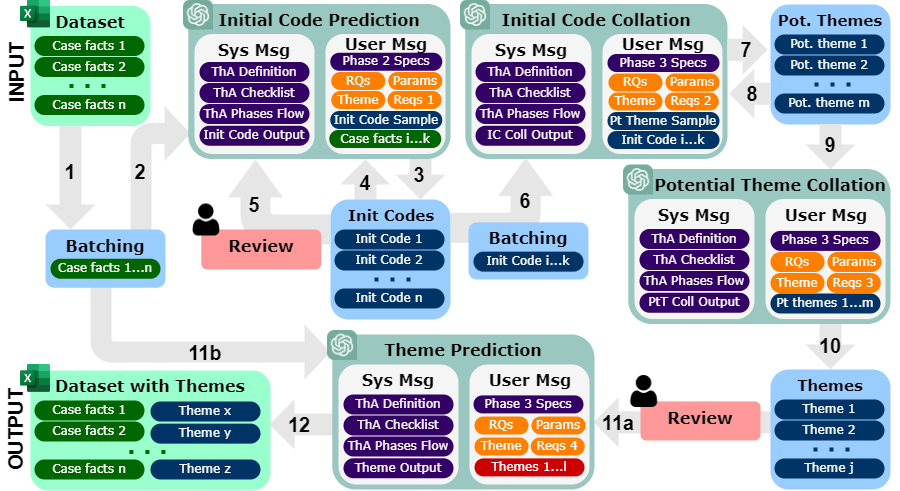}
    \caption{The diagram describes the proposed framework supporting phases 2 and 3 of thematic analysis. The data points are batched (1) to fit the LLM's prompt. Initial codes for each batch are predicted (2) in an iterative fashion utilizing the interim results (3 and 4). Expert feedback on the initial codes may be provided to trigger the re-generation of improved codes (5). The initial codes (batches) are collated into potential themes (6) in an iterative fashion utilizing the interim results (7 and 8). The potential themes are further collated into a compact list of high-level themes (9). The expert-reviewed themes (10) are predicted for the input data points (11a and 11b). The predictions of discovered themes for each data point are provided as the output (12).}
    \label{fig:framework}
\end{figure}

\paragraph{Model} The framework relies on OpenAI's GPT-4 model's capabilities  to perform complex NLP tasks in zero-shot settings \cite{openai2023gpt}. We set the \verb|temperature| of the model to 0.0, which corresponds to no randomness. Higher \verb|temperature| leads to more creative, but potentially less factual, output. We set \verb|max_tokens| to various values depending on the expected size of the output (a token roughly corresponds to a word). GPT-4 has an overall token length limit of 8,192 tokens, comprising both the prompt and the completion. We set \verb|top_p| to 1, as is recommended  when \verb|temperature| is set to 0.0. We set \verb|frequency_penalty| and  \verb|presence_penalty| to 0, which ensures no penalty is applied to repetitions and to tokens appearing multiple times in the output.

\paragraph{Resources} We utilize the definition of thematic analysis and the individual phases from~\cite{braun2006using}. For example, the 15-point checklist of criteria for good thematic analysis have been adopted verbatim as well as selected excerpts defining the analysis, its flow and the phases. Together with specifications of the expected outputs from various stages of the processing pipeline, these can be considered as \emph{general} resources, i.e., invariant to the performed thematic analysis. In addition, the framework requires \emph{context-specific} resources, i.e., different for each analysis. These include research questions, the parameters specifying the type of analysis (e.g., semantic/latent patterns, focus on a specific topic), the specification of what counts as a theme, and various sets of custom requirements.

\paragraph{Processing Flow} The proposed framework is depicted in Figure \ref{fig:framework}. The analyzed dataset is automatically segmented into batches where as many data points as can be fitted into the model's prompt are batched together, using the \verb|tiktoken| Python library.\footnote{Github: Tiktoken. Available at: \url{https://github.com/openai/tiktoken} [Accessed: 2023-04-30]} The data points are processed from the shortest to the longest. If a single data point exceeds the size of the prompt it is truncated to fit the limit by taking its starting and ending sequence~(both half the limit) and placing the ``\verb|[...]|'' token in between them. Each batch is inserted into the user message, alongside the research questions, other context-specific information about the analysis as well as any custom requirements. We also include a random sample of the initial codes predicted in the batches preceding the current one. The system message consists of the general resources. Using the \verb|openai| Python library,\footnote{GitHub: OpenAI Python Library. Available at: \url{https://github.com/openai/openai-python} [Accessed 2023-08-16]} the system and the user messages are then submitted to the LLM that generates a JSON file with the predicted initial codes. This is repeated until the whole dataset is labeled. The subject matter expert may review the predicted initial codes. Then, they may provide further custom instructions to the system as to what aspects of the data to focus on and which to disregard. These instructions are appended to the custom requirements. This process can be repeated until the predicted initial codes match the expectations.

The predicted initial codes are collated into potential themes. This stage of the processing is similar to the preceding one with the notable difference that the system operates on the batches of initial codes instead of the raw data points. The most common 20 potential themes predicted in the batches preceding the current one are included in the user message. As a result, each data point gets associated with a candidate theme. The candidate themes, which could be many, are then further collated into a compact set of high-level themes. The whole set of candidate themes is provided in the user message and submitted to the LLM. While this may depend on the specific analysis there is most likely no need to supply these in batches as all of them are likely to fit in the prompt. The output of this stage are the high-level themes (with candidate themes as sub-themes).

The final stage of the pipeline is focused on labeling the data points with the discovered themes. Note that this step differs from the prediction of the initial codes or potential themes because, here, the LLM is used to predict the labels from the provided list of themes (i.e., to perform deductive coding). The result of the whole process is the original data points being associated with (semi-)automatically discovered themes. This artifact can be utilized by the subject matter expert as a starting point for the subsequent phase of the thematic analysis (reviewing themes).

\section{Experimental Design}
\label{sec:experimental_design}

\paragraph{(RQ1) Autonomous Generation of Initial Codes} The quality of the automatically generated initial codes was manually assessed by one of the authors (a subject matter expert). The analysis suggested that, while largely sensible, the codes overly focused on \emph{what} was stolen, ignoring other aspects of the analysis (e.g., \emph{how} the offense was committed). To gauge the extent of the issue, we evaluated whether codes address (even implicitly) how the theft happened and what was stolen (evaluation scheme shown if Figure \ref{fig:ic_eval_scheme}). Each initial code was first analyzed with respect to \emph{$\neg$How}, and if the issue was confirmed the analysis stopped, i.e., the code was not further considered for \emph{$\neg$What}.

\begin{figure}
\begin{custombox}{Initial Codes Evaluation Scheme}
\begin{enumerate}
   \item \emph{$\neg$How}: If the code does not address (even implicitly) how the theft happened; $\rightarrow$
   \item \emph{$\neg$What}: If the code does not address (even implicitly) what was stolen; $\rightarrow$
   \item \emph{Ok}: The code addresses how the theft happened and what was stolen. $\square$
\end{enumerate}
\end{custombox}
   \caption{The coding scheme employed in evaluating the quality of the initial codes generated by the system autonomously (RQ1) and after the expert feedback was provided (RQ2).}
   \label{fig:ic_eval_scheme}
\end{figure}

\paragraph{(RQ2) Generating Initial Codes with Expert Feedback}
Following the analysis of the autonomously generated initial codes, we formulated compact instructions for the system to mitigate the most commonly appearing issues. The feedback consisted of two parts: (i) \emph{positive} (what to focus on) -- target, modus operandi, seriousness, and intent; and (ii) \emph{negative} (what to avoid) -- multiplicity of the offense, degree of completion, co-responsibility, value of stolen goods. We also provided three examples of desirable initial codes such as, e.g. ``\emph{vehicle theft with forceful entry and disassembly of vehicles}''. The instructions were included in the prompt (user message) as custom requirements. With the thus updated prompt, we repeated the generation of the initial codes. To assess the effects of the provided feedback, the newly generated codes were then manually coded using the same scheme as in the evaluation of RQ1, i.e., \emph{$\neg$How} $\rightarrow$ \emph{$\neg$What} $\rightarrow$ \emph{Ok} $\square$.

\paragraph{(RQ3) Predicting Themes}
To evaluate the zero-shot performance of the LLM in predicting themes for the analyzed data points, we employed the theme prediction component of the pipeline to label each data point with one of the themes arrived at by human experts. Note that a factual description may contain multiple themes, e.g., \textit{bicycle theft} and \textit{theft from an open-access place}, whereas the experts were instructed to assign the most specific and salient one. To account for this phenomenon, we instructed the system to also assign each data point with three of the themes. Then, we measured the performance of the system on this task in terms of recall at 1 (R@1) and recall at 3 (R@3).

\paragraph{(RQ4) Automatic Discovery and Prediction of Themes}
To investigate the performance of the proposed pipeline on the end-to-end task of autonomously discovering themes from the provided data, and assigning each analyzed data point with one of the identified themes, we employed the successive components of the pipeline to: (i) generate initial codes (with expert feedback); (ii) collate the initial codes into potential themes; (iii) group the potential themes into a compact list of higher-level themes; and (iv) assign each data point with one of the high-level themes. We then compared the automatically assigned themes to the manual themes discovered and assigned by subject matter experts.

\section{Results and Discussion}
\label{sec:results}
Table \ref{tab:results_init_codes} reports the results of the experiments focused on the prediction of initial codes. After the first round, 72.6\% of the 785 predicted codes were deemed reasonable, i.e., they described \emph{how} and \emph{what}. 13.2\% of the codes appeared to lack the focus on \emph{how}, and at least 14.1\% did not seem to describe \emph{what} was stolen. After the expert feedback was provided (Section \ref{sec:experimental_design}), 88.8\% of the codes were perceived as reasonable ($+16.2\%$ improvement). Table \ref{tab:compare_init_codes} compares example initial codes before and after the feedback, highlighting the improvements in coding the information of interest.\footnote{We admit a possible limitation of this experiment in that the author knew in which round the initial code was produced.} The results strongly suggest that the LLM can perform the initial coding of the data with reasonable quality~(RQ1), and further improve the codes upon receiving feedback from a subject matter expert~(RQ2). Hence, it appears that the proposed framework could become a valuable tool for supporting phase 2 of thematic analysis in ELS.

\begin{table}
    \centering
    \begin{tabular}{l|rr|rr}
        \toprule
        & \multicolumn{1}{c}{\emph{$\neg$How}} &\multicolumn{1}{c}{\emph{$\neg$What}}&\multicolumn{1}{|c}{\emph{Ok}}&\multicolumn{1}{c}{\emph{$\neg$Ok}}\\ 
        \midrule
        \textbf{Before expert feedback} & 104 (13.2\%) & 111 (14.1\%) &\cellcolor{positivehl}570 (72.6\%) &\cellcolor{negativehl}215 (27.4\%) \\
        \textbf{After expert feedback}  &  16 (2.0\%)  &  72 (9.2\%)  &\cellcolor{positivehl}697 (88.8\%) &\cellcolor{negativehl}88 (11.2\%) \\
        \bottomrule
    \end{tabular}
    \caption{Performance on the prediction of initial codes measured in terms of the evaluation scheme shown in Figure \ref{fig:ic_eval_scheme}. The inclusion of expert feedback into the prompt results in notable improvements across the board.}.
    \label{tab:results_init_codes}
\end{table}

\begin{table}
    \centering
    \footnotesize
    \setlength{\tabcolsep}{5pt}
    \begin{tabular}{p{\dimexpr 0.5\linewidth-2\tabcolsep}|p{\dimexpr 0.5\linewidth-2\tabcolsep}}
    \toprule
    \multicolumn{1}{c}{\bf Before expert feedback} & \multicolumn{1}{|c}{\bf After expert feedback}\\
    \midrule
         Private theft of cash \hln{from a residential space}	& Theft of large quantity of cash \hlp{from relative's home}\\
         Forced entry and \hln{theft} involving an automobile	& Burglary and \hlp{theft of work tools} from vehicle\\
         \hln{Shoplifting} - personal care items & Shoplifting of shaving equipment \hlp{from drugstore} \\
         \bottomrule
    \end{tabular}
    \caption{Examples of autonomously generated initial codes (left) and the initial codes generated after the subject matter expert feedback was provided (right). The colors highlight improvements (\hln{red} $\rightarrow$ \hlp{green}).}
    \label{tab:compare_init_codes}
\end{table}

The performance on the task of predicting themes (specified upfront) for the individual facts descriptions is described in Table \ref{tab:results_themes}. The prediction was performed using the list of 14 manually discovered themes (see Section~\ref{sec:dataset}). The overall R@1 of .66 and R@3 of .82 appear to suggest that the proposed approach is promising but clear limitations exist. This is largely consistent with prior related studies \cite{savelka2023unlocking,savelka2023can}. For some of the themes, e.g., \emph{theft in a shop} or \emph{theft of energy}, the automatic prediction worked remarkably well. However, there were also themes, e.g., \emph{theft from an open-access place} or \emph{robbing of cellars} where the performance was rather low. The promising results (RQ3) warrant investigations into the effects of providing expert feedback at this stage. This could either be done via providing additional custom instructions in the prompt and/or having the experts label a limited number of data points to be used in fine-tuning of the model.

\begin{table}
    \centering
    \setlength{\tabcolsep}{5pt}
    \begin{tabular}{lrr}
    \toprule
    Manual Theme          & R@1 & R@3 \\
    \midrule
    Theft in a shop                 & .95 & .96 \\
    Theft during a visit            & .75 & .91 \\ 
    Theft at work                   & .71 & .86 \\
    Breaking into another object    & .35 & .67 \\
    Pickpocketing                   & .68 & .87 \\
    Theft from an open-access place & .21 & .29 \\
    Theft of metals                 & .69 & .88 \\
    \bottomrule
    \end{tabular}
    \hspace{.4cm}
    \begin{tabular}{lrr}
    \toprule
    Manual Theme          & R@1 & R@3 \\
    \midrule
    Breaking into a residence       & .52 & .71 \\
    Bicycle theft                   & .74 & .96 \\ 
    Theft from a car                & .70 & .84 \\
    Robbing of cellars              & .14 & .57 \\
    Theft of motor vehicles         & .50 & .75 \\
    Theft of energy                 & 1.0 & 1.0 \\
    Others                          & .23 & .73 \\
    \bottomrule
    \end{tabular}
    \begin{tabular}{lrr}
    \bf Overall\hspace{2.75cm}      &\bf .66 &\bf .82 \\
    \bottomrule
    \end{tabular}
    \caption{Performance on the zero-shot prediction of themes discovered by subject matter experts from facts descriptions. R@1 is recall at 1 and R@3 is recall at 3.}
    \label{tab:results_themes}
\end{table}

\begin{figure}
    \includegraphics[width=\textwidth]{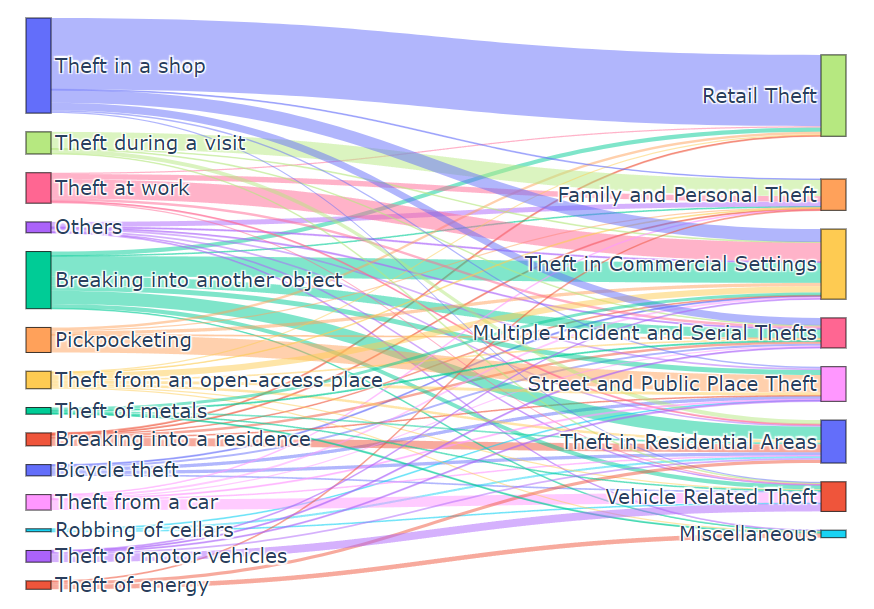}
    \caption{The graph shows mapping between themes discovered by subject matter expert (left) and the themes discovered by the proposed framework (right).}
    \label{fig:results_collation}
\end{figure}

The evaluation of the end-to-end performance in discovering and predicting themes~(RQ4) is shown in Figure \ref{fig:results_collation}. The number of themes discovered by the LLM (8) was less than the number of themes arrived at by the legal experts (14). For example, the LLM-discovered \emph{theft in commercial settings} maps to data points from manually discovered \emph{theft at work} and \emph{breaking into another object}. Some of the LLM-discovered themes appear to correspond to some of the manual ones (e.g., \emph{theft in a shop} $\rightarrow$ \emph{retail theft}). To further understand the mapping, we inspected the automatically generated potential themes from which the final 8 themes were automatically assembled. It appears that all of the manually identified themes could be mapped to one or more of the potential themes within a higher-level theme (e.g., \emph{robbing of cellars} $\rightarrow$ \emph{theft in residential areas}::\emph{burglary of storage area}). Further, the LLM identified behaviors that were missed, perhaps in error, by human experts (e.g., \emph{theft from gym}). 

Some of the potential themes were so similar that they should have been likely collapsed together. Other potential themes were overly specific (e.g., \emph{workplace theft} followed by various instances of what was stolen). Interestingly, the multiplicity of offending and/or stage of completion were present in some of the potential themes, despite specific instructions during the initial codes prediction not to focus on these aspects. Hence, an additional expert intervention in predicting potential themes might be warranted. The analysis strongly suggests that the LLM performs well in the end-to-end task of discovering and predicting themes from the raw data (RQ4). However, subject matter expert interventions might be desirable at various stages of the processing to improve the quality of the resulting themes and their alignment with the research questions. This echos with the cautioning sentiments expressed by De Paoli \cite{de2023can} and Jiang et al. \cite{jiang2021supporting} who reported that researchers performing qualitative analysis require full agency over the process. Moreover, the black-box nature of the proprietary LLMs is especially problematic from this point of view.

\section{Conclusions and Future Work}
We proposed a novel LLM-powered framework supporting thematic analysis, and evaluated its performance on an  analysis of criminal courts' opinions focused on the categories of thefts in Czechia. We found that the initial coding of data was performed with reasonable quality (RQ1), and further improved when expert feedback was provided~(RQ2). The performance on zero-shot classification of the data (facts descriptions) in terms of themes (categories of theft) was promising (RQ3) but could likely benefit from expert feedback (future work). The evaluation of the end-to-end performance of the pipeline on discovering and predicting themes suggested viability of the proposed framework (RQ4) while highlighting the importance of subject matter expert supervision. Besides incremental improvements, the future work should focus on extending the support beyond phases 2 and 3 of thematic analysis, and validating the findings of this study in other domains beyond court opinions and/or criminal law.

\bibliographystyle{vancouver}
\bibliography{main}
\end{document}